\pdfoutput=1

\documentclass[11pt]{article}

\usepackage[preprint]{acl}

\usepackage{times}
\usepackage{latexsym}
\usepackage{graphicx}
\usepackage{booktabs}
\usepackage{amsmath}
\usepackage{booktabs} 
\usepackage{tabularx} 
\usepackage{ragged2e} 
\usepackage{multirow} 
\usepackage{array} 
\usepackage{ltxtable}     
\usepackage{filecontents} 
\usepackage{longtable}
\usepackage[dvipsnames]{xcolor}
\usepackage{expex}
\usepackage{tikz}
\usetikzlibrary{calc} 
\usepackage{lineno}
\usepackage{graphicx}
\usepackage{svg}
\usepackage{float}

\usepackage[group-separator={,}]{siunitx} 

\usepackage{fontenc}

\usepackage[utf8]{inputenc}

\usepackage{microtype}

\usepackage{inconsolata}

\usepackage[dvipsnames]{xcolor} 

\newenvironment{xmlcode}
  {%
    \ttfamily%
  }
  {}

\newcommand{\xmltagfont}{\small}

\newcommand{\xmlopenN}[1]{%
  {\xmltagfont\textcolor{RoyalBlue}{<#1>}}%
}
\newcommand{\xmlcloseN}[1]{%
  {\xmltagfont\textcolor{RoyalBlue}{</#1>}}%
}

\newcommand{\xmlopenA}[2]{%
  {\xmltagfont
    \textcolor{RoyalBlue}{<#1}%
    \allowbreak\textcolor{BrickRed}{:#2}%
    \textcolor{RoyalBlue}{>}%
  }%
}
\newcommand{\xmlcloseA}[2]{%
  {\xmltagfont
    \textcolor{RoyalBlue}{</#1}%
    \allowbreak\textcolor{BrickRed}{:#2}%
    \textcolor{RoyalBlue}{>}%
  }%
}

\newcommand{\xmlopenB}[3]{%
  {\xmltagfont
    \textcolor{RoyalBlue}{<#1}%
    \allowbreak\textcolor{BrickRed}{:#2}%
    \allowbreak\textcolor{SeaGreen}{:#3}%
    \textcolor{RoyalBlue}{>}%
  }%
}
\newcommand{\xmlcloseB}[3]{%
  {\xmltagfont
    \textcolor{RoyalBlue}{</#1}%
    \allowbreak\textcolor{BrickRed}{:#2}%
    \allowbreak\textcolor{SeaGreen}{:#3}%
    \textcolor{RoyalBlue}{>}%
  }%
}

\title{\textit{You're Not Gonna Believe This}: A Computational Analysis of Factual Appeals and Sourcing in Partisan News}

\author{Guy Mor-Lan, Tamir Sheafer and Shaul R. Shenhav \\
        Hebrew University of Jerusalem \\
        \texttt{\{guy.mor, tamir.sheafer, shaul.shenhav\}@mail.huji.ac.il}}

\begin{document}
\maketitle
\begin{abstract}
While media bias is widely studied, the epistemic strategies behind factual reporting remain computationally underexplored. This paper analyzes these strategies through a large-scale comparison of CNN and Fox News. To isolate reporting style from topic selection, we employ an article matching strategy to compare reports on the same events and apply the \textsc{FactAppeal} framework to a corpus of over 470K articles covering two highly politicized periods: the COVID-19 pandemic and the Israel-Hamas war. We find that CNN's reporting contains more factual statements and is more likely to ground them in external sources. The outlets also exhibit sharply divergent sourcing patterns: CNN builds credibility by citing \texttt{Experts} and \texttt{Expert Documents}, constructing an appeal to formal authority, whereas Fox News favors \texttt{News Reports} and direct quotations. This work quantifies how partisan outlets use systematically different epistemic strategies to construct reality, adding a new dimension to the study of media bias.
\end{abstract}

\section{Introduction}

In a fragmented and polarized media ecosystem, the question of how news outlets present reality has profound implications for democratic discourse. Partisan news sources do not merely offer different opinions; they often construct divergent factual narratives around the same events, creating distinct ``epistemic worlds'' for their audiences \cite{iyengar2009red, broockman2025consuming}. This divergence is achieved not only through topic selection—what has been termed ``partisan coverage filtering'' \cite{broockman2025consuming}—but also through subtle linguistic choices in how information is framed \cite{entman1993framing}, which facts are presented, and crucially, which sources are invoked to lend those facts authority.

This study focuses on CNN and Fox News, two of the most influential news organizations in the United States, both of which have been shown to operate as partisan outlets. While foundational research on their television broadcasts has demonstrated their power to causally shift political preferences and impact election outcomes \cite{ash2024viewers}, their online news platforms are formidable forces in their own right. According to recent industry analyses, CNN.com and FoxNews.com consistently rank among the most visited news websites in the US, surpassed only by legacy outlets like The New York Times \cite{prlab2025topmedia}. Prior work has powerfully documented the effects of ``partisan coverage filtering'' on television—the strategic selection and omission of facts—which shapes audience beliefs and attitudes \cite{broockman2025consuming}. However, these studies primarily reveal \textit{what} topics are covered differently. The more granular question of \textit{how} these outlets rhetorically construct factual authority within their reporting remains computationally underexplored. This study, therefore, moves beyond topic selection to provide the first large-scale computational analysis of the epistemic and sourcing strategies used in their widely-read digital articles, offering a new lens to understand the mechanics of partisan media in the digital age.

How is a factual claim made credible? A news report can present a ``brute fact'' on its own authority, or it can anchor the claim in an external source of knowledge—an official, an expert, a witness, or a document. This choice of whether to appeal to a source, and which type of source to appeal to, is a central journalistic practice \cite{schudson2003sociology} that shapes the perceived legitimacy of information \cite{carlson2017journalistic}. The differential use of these \textbf{epistemic appeals} may represent a powerful, yet computationally understudied, form of media bias.

In this work, we conduct a large-scale computational analysis of epistemic and sourcing strategies in partisan news. We investigate whether right-leaning (Fox News) and left-leaning (CNN) media outlets exhibit systematic differences in (1) the proportion of factual versus non-factual statements; (2) the tendency to support factual claims with explicit epistemic appeals; and (3) the distribution of source types (e.g., Officials, Experts) used to validate claims.

To address these questions, we leverage the \textit{FactAppeal} annotation framework \cite{mor-lan-2024-factappeal}, a detailed scheme for identifying factual claims and the structure of their epistemic appeals. We apply a model trained on this scheme to a new, curated corpus of news articles from CNN and Fox News. To isolate stylistic differences from topic selection bias, we use an article matching strategy to compare reports covering the same events during two highly polarized periods: the COVID-19 pandemic and the Israel-Hamas war.

Our findings reveal significant and systematic differences in the epistemic postures of the two outlets. We show that while both outlets are predominantly factual, they diverge sharply in how they source their claims, with Fox News prioritizing appeals to other media and direct quotes, while CNN prioritizes appeals to formal expertise. This research contributes a novel, quantitative methodology for analyzing sourcing as a vector of media bias and provides empirical evidence of the distinct ways partisan outlets build factual authority.

\section{Related Work}

This study is situated at the intersection of computational linguistics, political communication, and journalism studies. We build on research in three primary areas: computational analysis of media bias, the study of journalistic sourcing, and NLP for epistemic analysis.

\subsection{Computational Analysis of Media Bias}
The computational study of media bias has a rich history, moving from lexical choices \cite{gentzkow2010what} to more nuanced phenomena like framing \cite{entman1993framing, card2015media}. A key mechanism of framing in modern partisan media is ``partisan coverage filtering," where outlets selectively report facts favorable to their side \cite{broockman2025consuming}. Recent work has also used unsupervised methods combining named entity recognition and stance analysis to cluster outlets based on topic and portrayal patterns \cite{benson2023developing, deVries26102022}.

Our work extends this tradition by proposing that epistemic structure—the very manner in which a factual claim is presented and supported—is itself a powerful framing device. Instead of focusing only on \textit{what} is said or filtered, we analyze \textit{how} the remaining claims are made credible. This moves beyond lexical or topic bias to what can be termed ``epistemic bias'' or ``source bias.''

\subsection{Journalistic Sourcing and Authority}

Sociologists of news have long argued that sourcing practices are central to the construction of news narratives \cite{schudson2003sociology}, often leading journalists to "index" their coverage to the range of opinions found among elite political sources \cite{bennett1990toward}. The choice of sources is not neutral; it confers legitimacy on certain viewpoints while marginalizing others \cite{carlson2017journalistic}. This can lead to phenomena like ``false balance," where giving equal weight to unequal sides creates a distorted perception of evidence, a form of epistemic harm \cite{boykoff2004balance, terzian2025epistemic}.

In a partisan context, source selection is also a persuasive tool. Citing a source already trusted by a particular audience segment can increase a report's persuasive power for that segment \cite{iyengar2009red}. Our work provides a computational method to systematically measure these sourcing patterns at scale, operationalizing them as a quantifiable vector of media influence. While some studies have found that sustained exposure to cross-cutting media can moderate attitudes despite source distrust \cite{broockman2025consuming}, others find that it primarily erodes trust in all media \cite{guess2021consequences}, highlighting the complex role of sourcing.

This sociological perspective can be enriched by the framework of social epistemology, which examines the social practices that facilitate or hinder the creation of knowledge \cite{goldman1999knowledge}. From this view, journalistic sourcing is not just a persuasive tool but a core epistemic practice. The systematic preference for certain source types over others can lead to what philosopher Miranda Fricker terms ``testimonial injustice,'' where the credibility of certain knowers is unfairly diminished \cite{fricker2007epistemic}. Furthermore, as argued by C. Thi Nguyen, such practices are central to the distinction between an ``epistemic bubble'' (a simple lack of information) and a more pernicious ``echo chamber,'' which actively works to discredit outside sources of authority \cite{nguyen2020echo}.

\subsection{Epistemic Analysis in NLP}
Within NLP, our research builds on work in claim detection, fact-checking, and argumentation mining. Large-scale datasets like FEVER \cite{thorne2018fever} have driven progress in verifying claims against evidence, and argumentation mining seeks to identify claim-premise structures \cite{feng2011classifying}. However, these tasks typically do not provide a detailed typology of the \textit{kind} of evidence or authority being invoked, nor do they capture nuanced factual relationships \citep{mor-lan-levi-2024-exploring}. Similarly, while stance detection identifies a text's position on a target, it does not analyze the evidentiary basis of factual claims \cite{kucuk2020survey, hardalov2022survey}.

The \textit{FactAppeal} framework \cite{mor-lan-2024-factappeal}, upon which our study is based, was designed to fill this gap. It provides a rich, theory-driven schema for not just identifying a factual claim, but for detailing how it is epistemically grounded via appeals to different kinds of sources. By applying this fine-grained analytical lens to a partisan news corpus, we move from general claim detection to a nuanced analysis of how partisan arguments are constructed and justified.

\section{Methodology}

\begin{figure*}[h!]
    \centering
    \includegraphics[width=1.0\textwidth]{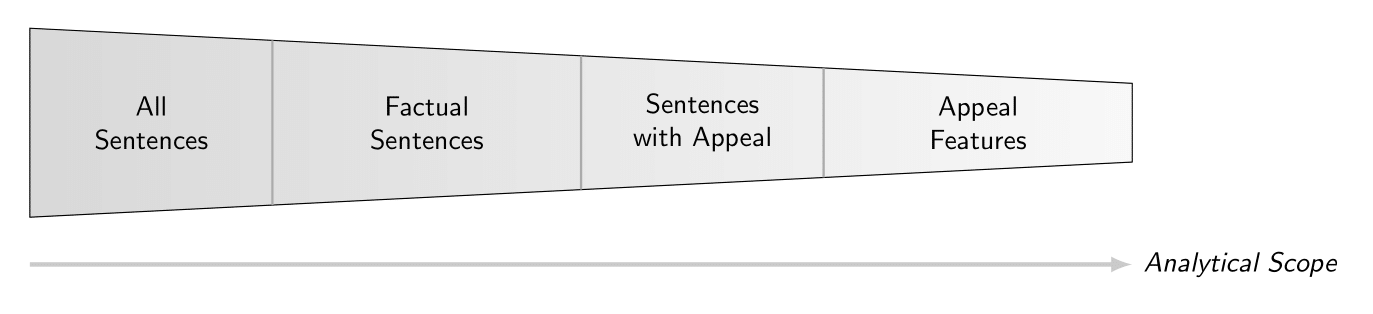}
    \caption{Analytical Funnel}
    \label{fig:funnel}
\end{figure*}

Our analytical approach is structured as a conceptual funnel (Figure~\ref{fig:funnel}), which allows us to systematically compare how journalistic outlets construct factual claims. The analysis begins at the broadest level—all sentences within a text—and progressively narrows its scope. We first identify the subset of sentences that make factual claims, then the subset of those that ground their claims in an epistemic appeal, and finally, we analyze the fine-grained characteristics of these appeals, such as source type and quotation style.

\subsection{The FactAppeal Framework}
Our analysis relies on the \textit{FactAppeal} annotation scheme, a span-level framework for identifying the epistemic structure of factual claims in news text \cite{mor-lan-2024-factappeal}. The scheme first distinguishes sentences that make \textbf{factual claims}—statements about the state of the external world—from non-factual content. Factual claims are then categorized as either a \textbf{Fact Without Appeal} (a brute fact presented on the author's authority) or a \textbf{Fact With Appeal}.

For sentences containing a \texttt{Fact With Appeal}, the framework annotates the \texttt{Source} of the appeal and classifies it into a typology grounded in the nature of its authority \cite{mor-lan-2024-factappeal}. Key source types include:
\begin{itemize}
    \item \textbf{Official:} An actor with non-epistemic authority (e.g., political, legal) over the events.
    \item \textbf{Expert:} A source whose authority derives from specialized, general knowledge (e.g., a scientist, an analyst).
    \item \textbf{Active Participant:} An actor with a direct, firsthand role in the events.
    \item \textbf{Witness:} An observer with firsthand testimony but not an active role.
    \item \textbf{Direct Evidence:} Non-human evidence from the scene (e.g., a photo, a document).
    \item \textbf{News Report:} An appeal to a journalistic news report.
    \item \textbf{Expert Document:} An appeal to a formal, expert document (e.g., a scientific study, an institutional report).
\end{itemize}
The scheme also annotates other features, such as whether the source is explicitly named, whether the appeal involves a direct or indirect quote, and other characteristics such as attributes of the source (role, title or credentials) or of the epistemic appeal, such as the time or location in which the appeal was made \cite{mor-lan-2024-factappeal}.

To illustrate, the following example shows how a sentence is annotated with XML-style tags to capture its epistemic structure.

\exdisplay\noexno
\footnotesize
\begin{xmlcode}
\xmlopenB{Source}{Named}{Official} Doug Ericksen, \xmlcloseN{Source} \xmlopenN{Source\_Attribute} the EPA's communications director for the transition, \xmlcloseN{Source\_Attribute} told \xmlopenN{Recipient} National Public Radio \xmlcloseN{Recipient} that \xmlopenA{Fact\_Appeal}{Direct} ``we'll take a look at what's happening so that the voice coming from the EPA is one that's going to reflect the new administration." \xmlcloseN{Fact\_Appeal}
\end{xmlcode}
\xe

This fine-grained annotation allows us to move beyond simple claim detection to a nuanced analysis of how factual claims are constructed and justified.

\subsection{Data Collection}
We gathered a corpus of news articles from the websites of CNN (left-leaning) and Fox News (right-leaning) via the Webz.io API. The data covers two broad periods: February 2020 to June 2022 and September 2023 to June 2025. The final annotated corpus consists of 476K articles, comprising 16.5M sentences. Table \ref{tab:desc_stats} provides a summary.\footnote{The $2^\text{nd}$ period contains considerably more video-only items, which are filtered.}

\begin{table}[h!]
\centering
\resizebox{\columnwidth}{!}{%
\begin{tabular}{@{} >{\RaggedRight}p{4cm} l S[table-format=6.0] S[table-format=7.0] @{}}
\toprule
\textbf{Period} & \textbf{Source} & {\textbf{Articles}} & {\textbf{Sentences}} \\
\midrule
\multirow{2}{*}{Feb 2020 -- Jun 2022} & CNN & 194424 & 7549323 \\
& Fox News & 176581 & 5502859 \\
\midrule
\multirow{2}{*}{Sep 2023 -- Jun 2025} & CNN & 45814 & 1567906 \\
& Fox News & 59814 & 1837709 \\
\bottomrule
\end{tabular}%
}
\caption{Descriptive Statistics of the Dataset}
\label{tab:desc_stats}
\end{table}

\subsection{Article Matching}

Differences between news outlets can arise from their editorial choices about \textit{what} stories to cover, or \textit{how} they frame and report on the same story. Topic selection acts as a powerful confounding variable: a simple comparison between outlets might incorrectly attribute differences in reporting style to the outlets themselves, when they are actually due to the different events and stories each chooses to cover, which may lend themselves to different sourcing patterns. For example, Lifestyle articles might naturally contain fewer epistemic appeals and fewer appeals to experts. To isolate the effect of reporting style from this topic-selection bias, we employ a matching strategy analogous to methods used in statistics to control for confounders. By pairing articles that cover the same event, we can more directly attribute any observed differences to the framing and journalistic practices of the outlets. This approach allows us to approximate a controlled comparison, conditioning our analysis on the underlying story.

Our matching procedure operates daily. For each day $t$ in our corpus, let $A_t$ and $B_t$ represent the sets of articles from the two outlets. We first represent each article's title as a vector embedding, $\mathbf{v}$, using OpenAI's \texttt{text-embedding-3-small} model. Then, for each article $a \in A_t$, we calculate its maximum cosine similarity against the set of articles published by the opposing outlet on the same day. This yields a similarity score for article $a$, denoted $Sim(a)$, defined as:

$$
Sim(a) = \max_{b \in B_t} \left( \frac{\mathbf{v}_a \cdot \mathbf{v}_b}{\|\mathbf{v}_a\| \|\mathbf{v}_b\|} \right)
$$

This process is applied symmetrically to assign a similarity score to every article from both outlets. Each article in the corpus is thus associated with a score representing its strongest topical link to an article from the other source on the day of its publication.

As shown by the stratified random sample of matched titles in Table~\ref{tab:matched_titles_sorted}, this score effectively captures topical alignment. High-similarity scores correspond to articles covering the same specific event, while medium scores indicate a shared general topic. Based on this, we partition the entire corpus of articles $C = A \cup B$ into three disjoint analytical groups: the \textbf{High-Similarity} group, $G_{high} = \{ a \in C \mid Sim(a) \ge 0.6 \}$; the \textbf{Mid-Similarity} group, $G_{mid} = \{ a \in C \mid 0.3 < Sim(a) \le 0.6 \}$; and the \textbf{Low-Similarity} group, $G_{low} = \{ a \in C \mid Sim(a) \le 0.3 \}$.

\begin{figure}[h!]
    \centering
    \includegraphics[width=0.5\textwidth]{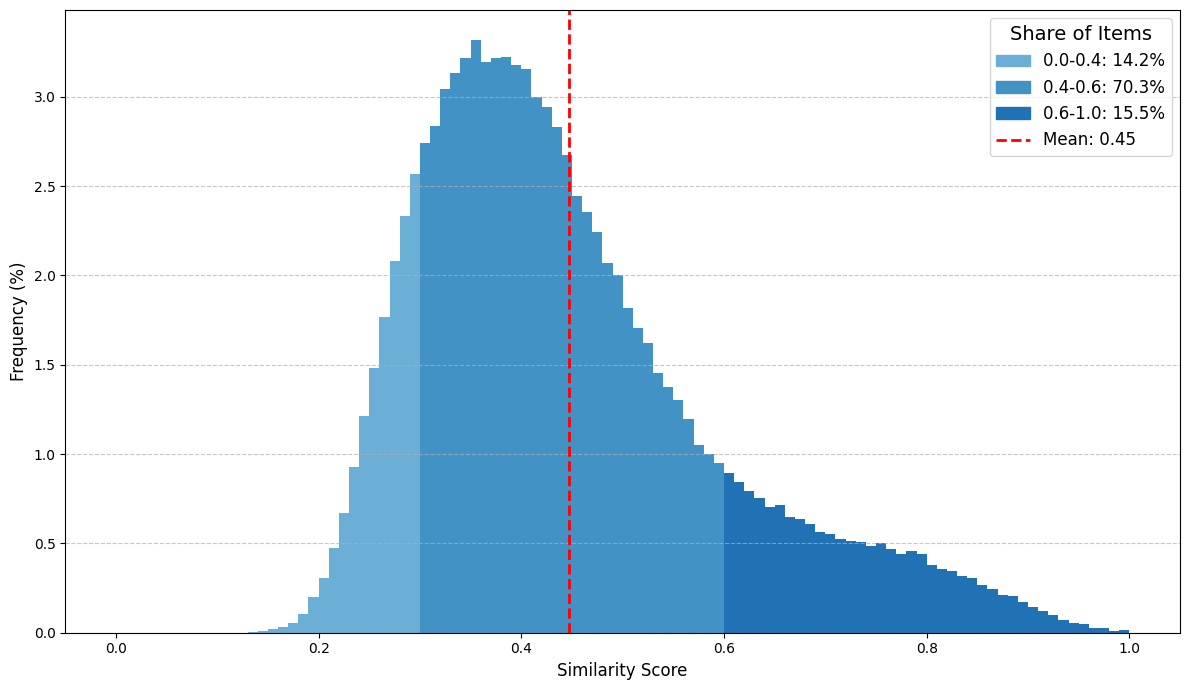}
    \caption{Histogram of Similarity Scores}
    \label{fig:sim_hist}
\end{figure}

Figure~\ref{fig:sim_hist} shows the distribution of these similarity scores and the resulting proportion of articles in each group. We sample 100 pairs of titles from $G_{high}$ for manual validation, and observe that all matched articles report on the same event or story. For additional examples of matched titles in the high similarity group, see Table~\ref{tab:matched_titles} in the appendix. This tiered approach allows us to analyze differences in reporting practices while controlling for the degree of topical overlap between the articles.

\begin{table*}[htbp]
    \centering
    \footnotesize 
    \begin{tabularx}{\textwidth}{ >{\RaggedRight}X >{\RaggedRight}X c }
        \toprule
        \textbf{CNN Title} & \textbf{Fox News Title} & \textbf{Similarity} \\
        \midrule
        Art Howe: Former MLB player and manager is in the ICU battling coronavirus, report says - CNN & Art Howe, former MLB manager and infielder, battling coronavirus in ICU & 0.92 \\
        \addlinespace
        Arthur Ochs Sulzberger, Jr., to retire as NY Times chairman & NY Times chairman Arthur Sulzberger Jr. retiring, handing role to son & 0.81 \\
        \addlinespace
        Nissan unveils its first electric SUV, the Ariya & Electric Nissan Ariya to take on Tesla Model Y with 300-mile range, \$40G price & 0.73 \\
        \addlinespace
        Supreme Court: John Roberts sides with liberals to block controversial Louisiana abortion law - CNNPolitics & Supreme Court strikes down Louisiana abortion restrictions & 0.68 \\
        \addlinespace
        American teacher who taught Ukrainian mom English raises funds to help her flee & Ukrainian American school custodian in GA stunned by student artwork honoring Ukraine & 0.52 \\
        \addlinespace
        Trump is undermining post office to increase his reelection chances (opinion) - CNN & NJ Gov. Murphy 'unequivocally' supports more funding for postal system after mail-in voting announcement & 0.49 \\
        \addlinespace
        The pandemic stock market divide isn't going away & Psaki hints Biden may skip bipartisan deal on COVID-19 if needed: 'Not going to take any tools off the table' & 0.36 \\
        \addlinespace
        NFL updates Covid-19 protocols and adds video monitoring of teams - CNN & Relieved Cincinnati Bengals now look for momentum & 0.30 \\
        \addlinespace
        This course bundle covers more than we knew we wanted to learn about Excel & What to know on Election Day 2020: Culmination of the battle for the White House & 0.20 \\
        \addlinespace
        Transforming an A330 into a yacht-like luxury experience & Lightning beat Rangers 2-1, advance to Stanley Cup Final & 0.10 \\
        \bottomrule
    \end{tabularx}
    \caption{Examples of Matched CNN and Fox News Article Titles with Cosine Similarity}
    \label{tab:matched_titles_sorted}
\end{table*}

\subsection{Model Inference}

We segment each article into sentences using the SpaCy segmentizer model. We then process each sentence using \textsc{FactAppeal}'s best performing checkpoint, a fine-tuned Gemma-2-9B decoder, achieving a macro-$F_1$ score of 0.73 \citep{mor-lan-2024-factappeal}. The trained model is prompted to reproduce each input sentence with XML-style tags inserted to indicate features of epistemic appeals. Table~\ref{tab:annotated_comparison} in the appendix shows a side-by-side comparison of two example articles annotated by the model. Inference is performed on an A100 GPU using the vLLM framework. Following inference, we parse the tags and create counts of each feature.

\section{Analysis and Results}
\label{sec:results}
Our analysis reveals systematic differences in the epistemic strategies employed by CNN and Fox News. We examine these differences by calculating several conditional probabilities, comparing the outlets across our three similarity groups to control for topic selection effects.

\subsection{Factuality and Propensity to Appeal}
\begin{figure}[htbp]
\centering
\def\svgwidth{0.5\textwidth}
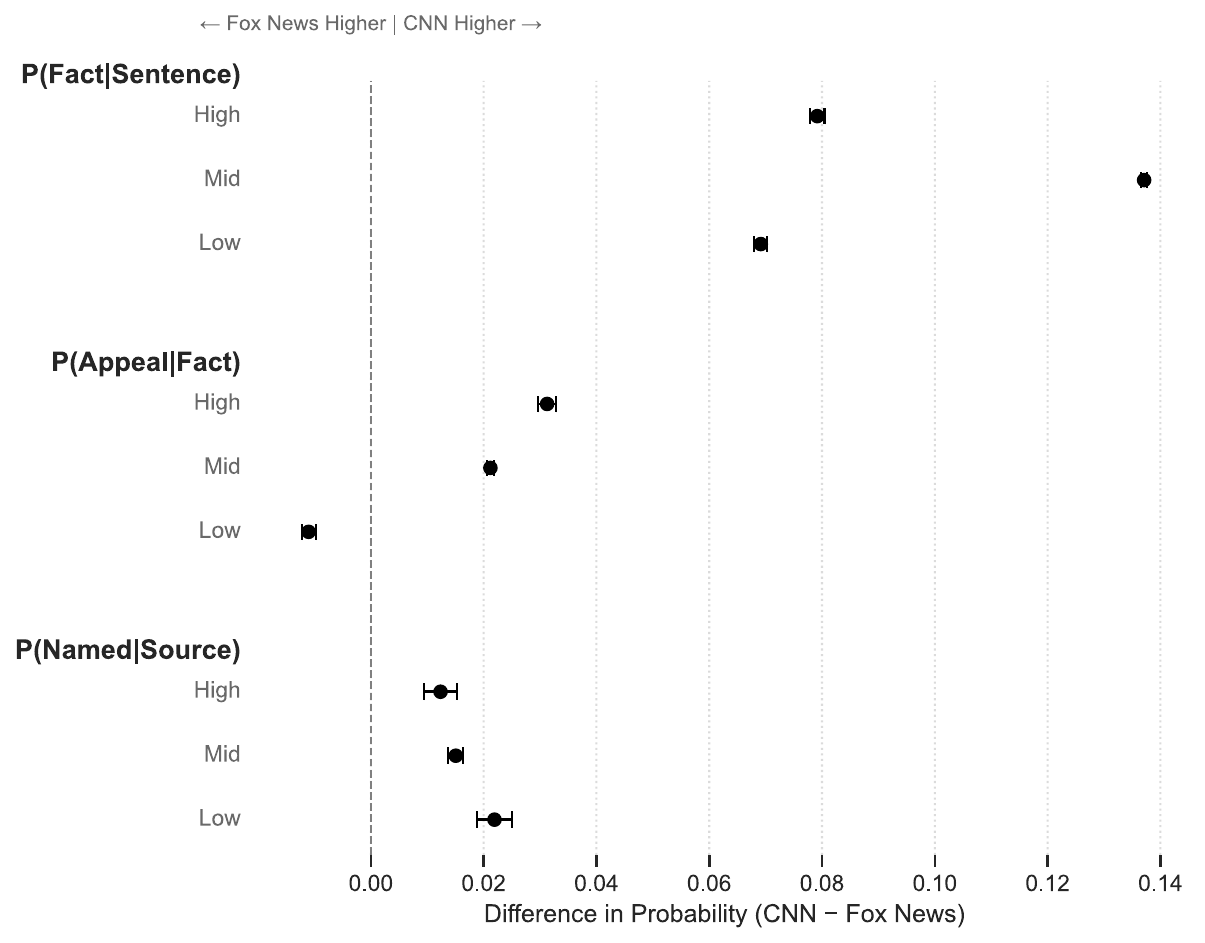
\caption{Difference in Factuality and Appeals}
\label{fig:key_probs}
\end{figure}

We first measure the outlets' general journalistic posture: the tendency to report facts and to ground those facts in external sources. Figure \ref{fig:key_probs} shows the difference between CNN and Fox News in three key probabilities:
\begin{itemize}
    \item $P(\text{Fact}|\text{Sentence})$: The probability that a given sentence contains a factual claim.
    \item $P(\text{Appeal}|\text{Fact})$: The probability that a factual claim is supported by an epistemic appeal.
    \item $P(\text{Named}|\text{Source})$: The probability that a cited source is identified by name.
\end{itemize}

Our results show a clear and consistent pattern. Sentences in CNN articles are significantly more likely to express factual statements across all similarity groups. Furthermore, when a factual statement is made, CNN is more likely to accompany it with an epistemic appeal, except in the low-similarity group where the finding is reversed. Finally, CNN is also more likely to identify its sources by name.

These findings suggest a fundamental difference in journalistic style. CNN's reporting exhibits a greater orientation toward fact-based statements that are explicitly anchored to named, external authorities. This aligns with a traditional journalistic norm of transparently sourced, evidence-based reporting.

\subsection{Divergent Sourcing Strategies}
\begin{figure}[htbp]
\centering
\def\svgwidth{0.5\textwidth}
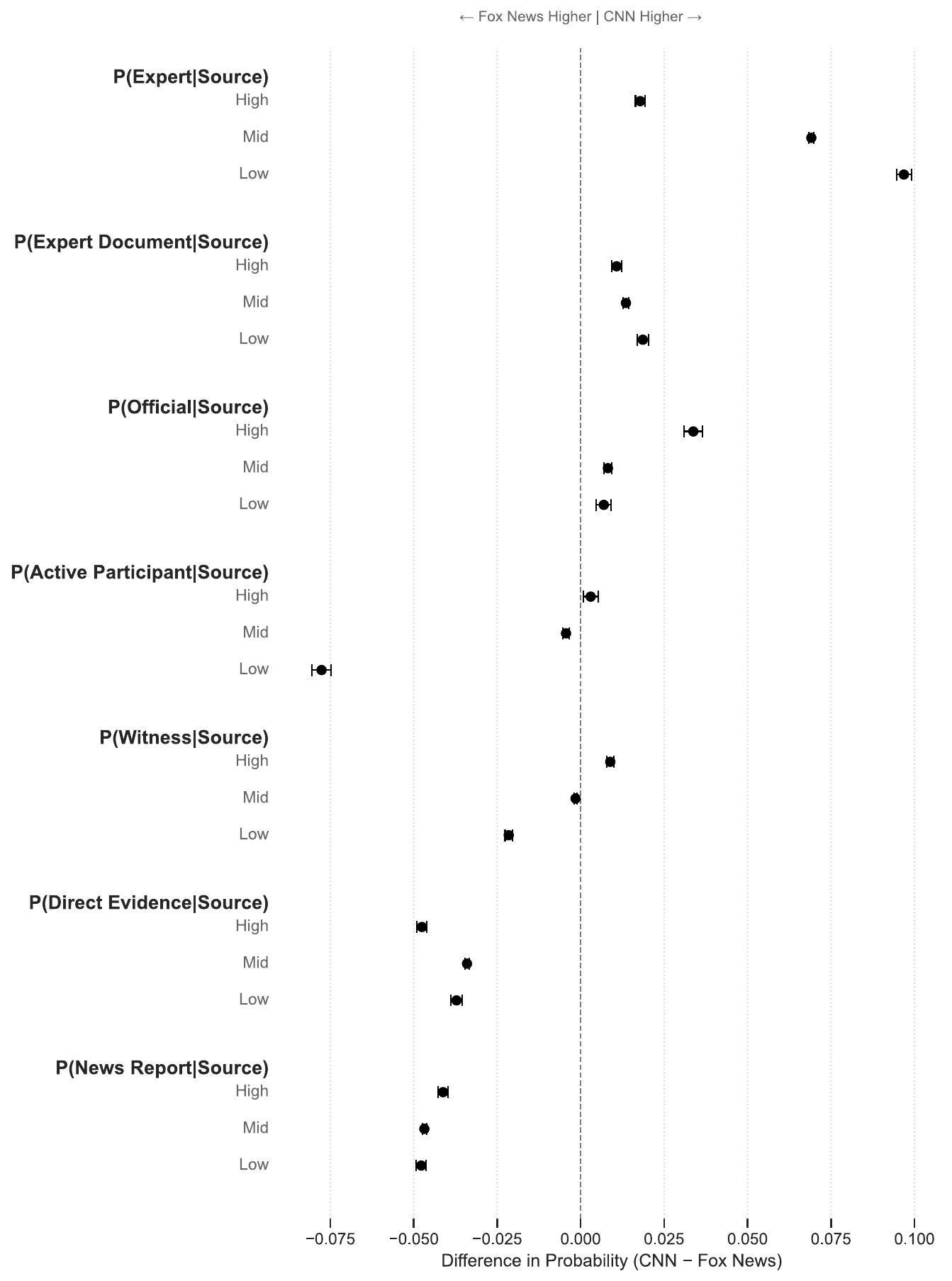
\caption{Difference in Appeal Source Types}
\label{fig:source_types}
\end{figure}

Beyond the frequency of appeals, the \textit{type} of source invoked reveals deeper strategic differences. Figure \ref{fig:source_types} shows the difference in the conditional probability of citing each source type, $P(\text{Type}=t|\text{Source})$.

Here, the two outlets display starkly divergent patterns. CNN consistently shows a significantly higher propensity to cite \texttt{Experts}, \texttt{Expert Documents}, and \texttt{Officials}. This pattern suggests a strategy of building credibility through appeals to formal, institutional, and scientific authority.

Conversely, Fox News relies significantly more on \texttt{News Reports} and \texttt{Direct Evidence}. The reliance on other news reports may indicate a greater role in inter-media agenda-setting or the creation of a self-referential media ecosystem. The preference for \texttt{Direct Evidence} suggests an appeal to unmediated, ``raw'' proof, perhaps reflecting a degree of skepticism toward expert- or official-based sources. Again, the occurrence of these differences in the high-similarity group, where both outlets cover the same story, indicates this is a framing choice rather than a byproduct of topic selection.

\subsection{Characteristics of Appeals}
\begin{figure}[htbp]
\centering
\def\svgwidth{0.5\textwidth}
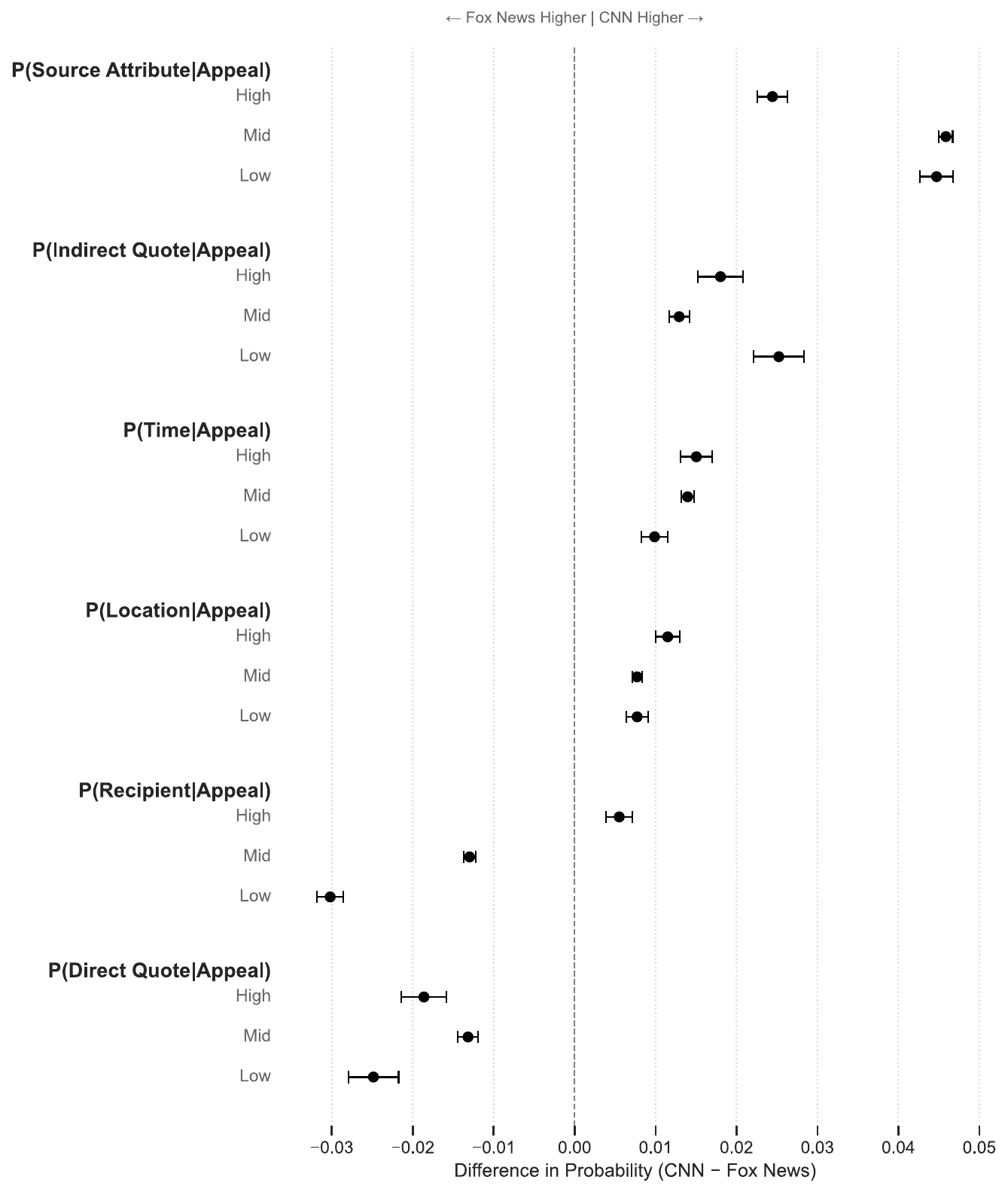
\caption{Difference in Appeal Characteristics}
\label{fig:appeal_features}
\end{figure}

Finally, we examine finer-grained features of the epistemic appeals themselves (Figure \ref{fig:appeal_features}). CNN is more likely to report the \texttt{Source Attribute} (e.g., role or credentials) and the \texttt{Time} and \texttt{Location} of the appeal, providing richer contextual information that further serves to credentialize the source.

CNN is also more likely to utilize indirect quotations, in which the epistemic source's words are paraphrased and mediated. In contrast, Fox News is significantly more likely to use a \texttt{Direct\_Quote}. This preference for verbatim quotation, combined with its lower rate of providing source attributes, suggests a strategy that prioritizes conveying a source's message with seemingly unfiltered immediacy, rather than contextualizing their authority.

\subsection{Topic and Expertise Analysis}
\begin{figure*}[!h]
\centering
\def\svgwidth{1.0\textwidth}
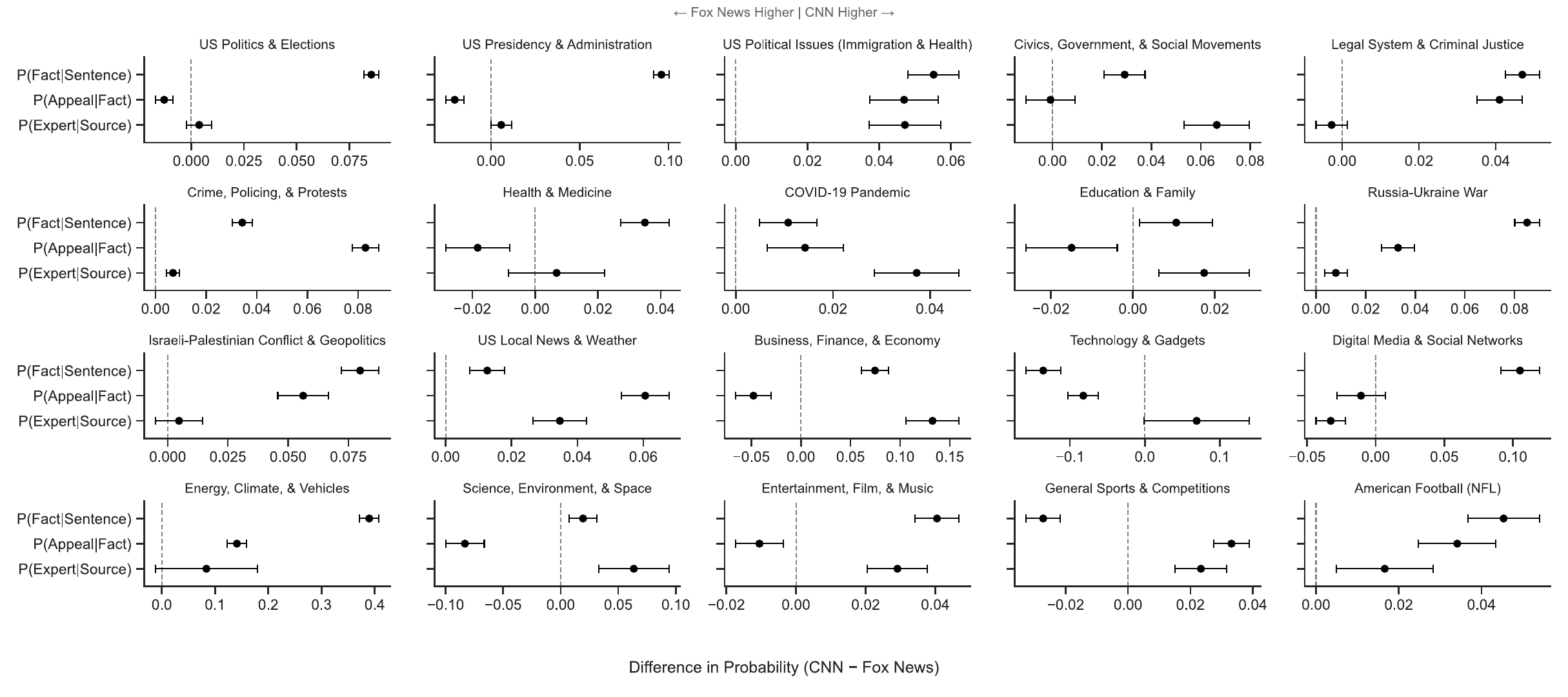
\caption{Main Differences by Topic...}
\label{fig:by_topic}
\end{figure*}

To test the robustness of these findings, we conducted a topic-level analysis using an LDA topic model \cite{blei2003latent}. As shown in Figure \ref{fig:by_topic}, the higher factuality rate for CNN ($P(\text{Fact}|\text{Sentence})$) is consistent across nearly all topics. However, the differences in the propensity to make appeals and to cite experts vary by topic, suggesting that sourcing strategies are also context-dependent.

We observe the greatest disparity in reporting practices across topics that are highly politically polarizing yet fundamentally reliant on substantive factual claims (i.e., non-normative dimensions). These topics include immigration and health, the COVID-19 pandemic, climate and energy, and crime and policing. The heightened differences in these areas indicate that the choice of epistemic strategy is actively leveraged to construct distinct narratives around politically charged facts.

\begin{figure*}[!h]
\centering
\def\svgwidth{1.0\textwidth}
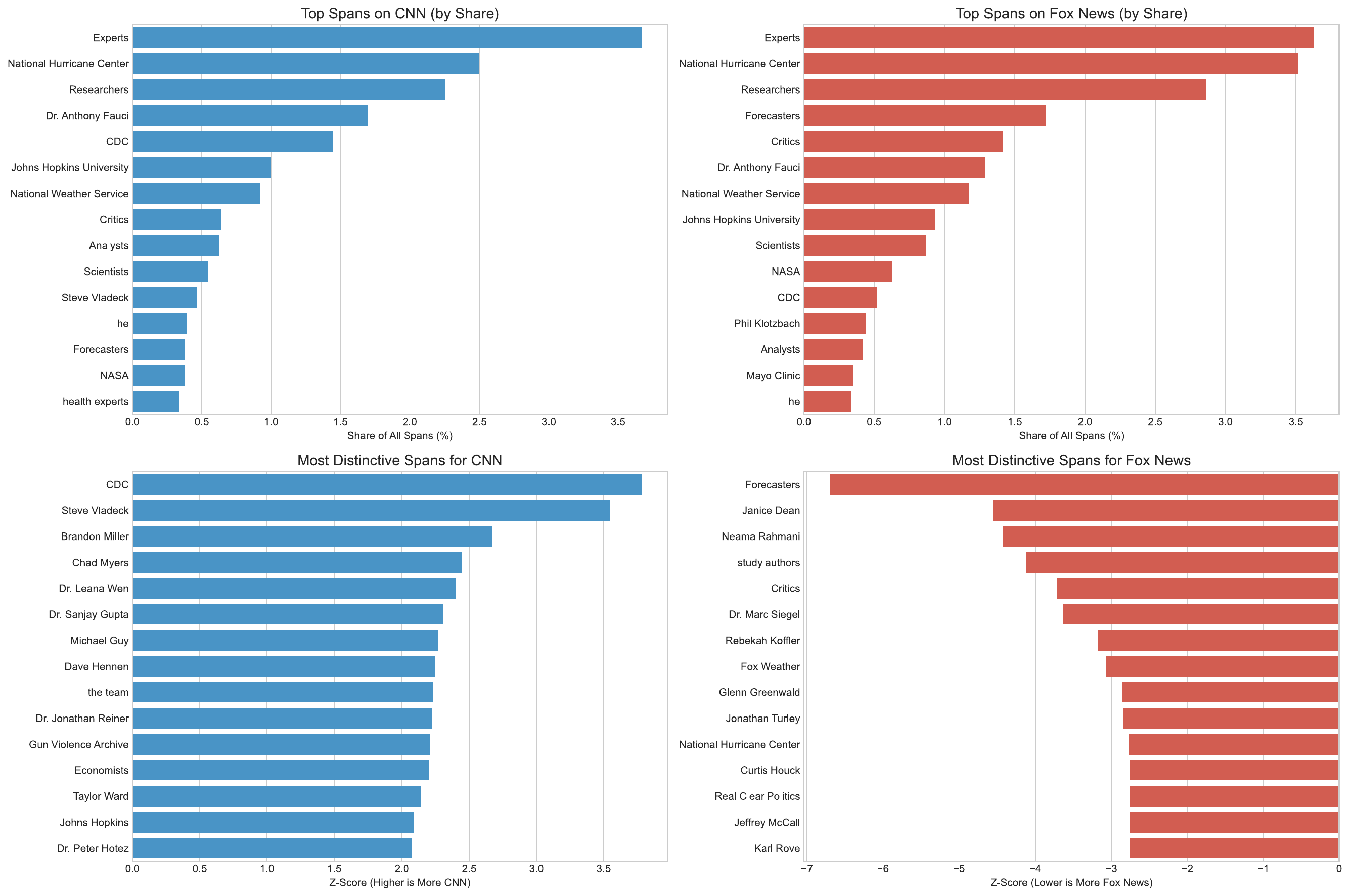
\caption{Most Frequent and Distinctive Expert Spans...}
\label{fig:expert_spans}
\end{figure*}

Given that the preference for \texttt{Expert} sources is one of the most pronounced differences, we performed a deeper analysis of the specific expert spans cited by each outlet. Figure \ref{fig:expert_spans} displays the most common expert spans cited by each outlets, as well as the most distinctive spans, which we calculate using weighted log-odds \cite{Monroe2008}. While the top expert spans are generally shared on the two venues, their ordering and relative shares differ. The CDC is the most distinctive expert source for CNN, reflecting its focus on public health expertise during the pandemic. Both outlets rely on their own in-house experts (e.g., Steve Vladeck for CNN, Dr. Marc Siegel for Fox News), reinforcing their respective authority networks. Notably, one of the most distinctive expert-like sources for Fox News is ``Critics,'' suggesting a strategy of framing issues through an oppositional lens, even when the source of criticism is not a traditional expert.

\section{Discussion}

The results paint a clear picture of divergent journalistic epistemologies in partisan news. While both CNN and Fox News heavily engage in factual reporting, the way they construct the authority for those facts differs systematically.

CNN's approach can be characterized as an appeal to \textit{formal, institutional authority}. Its higher rate of factuality and appeals, its preference for experts and official documents, and its tendency to provide source credentials all work to project an image of objective, evidence-based journalism. This strategy aligns with traditional journalistic norms and may appeal to an audience that values institutional credibility.

Fox News, in contrast, employs a strategy that can be described as an appeal to \textit{media-centric and unmediated authority}. Its greater reliance on other news reports creates a more self-referential information environment, while its preference for direct quotes and direct evidence creates a sense of immediacy and raw, unfiltered truth. The frequent citation of ``critics'' as an authority further frames news as a landscape of conflict.

These distinct sourcing patterns contribute to the creation of separate epistemic worlds for their audiences. Audiences are not only exposed to different facts via partisan coverage filtering \cite{broockman2025consuming}, but are also socialized into different models of what constitutes a valid source of knowledge. Over time, this can lead to a situation where a claim's credibility is judged not on its own merits, but on whether it comes from an ``approved'' type of source for one's political tribe. This ``epistemic skewing'' deepens polarization by eroding the common ground for factual debate, making cross-partisan communication increasingly difficult. These distinct sourcing patterns are the very mechanism by which epistemic ``echo chambers'' are constructed and maintained \cite{nguyen2020echo}. By consistently elevating certain types of sources while implicitly downplaying others, partisan outlets instruct their audiences not just on \textit{what} to believe, but on \textit{who} to believe, thereby inoculating them against countervailing evidence. 

\section{Conclusion}
In this paper, we introduced a novel computational approach to analyze sourcing and epistemic appeals as a dimension of media bias. By applying a fine-grained classification model to a matched corpus of articles from CNN and Fox News, we demonstrated that these outlets employ systematically different strategies to ground their factual claims. Our findings quantify how partisan media outlets construct different authoritative realities for their audiences, moving beyond lexical or topic bias to reveal a deeper, structural ``epistemic bias.'' This work opens new avenues for the large-scale, empirical study of journalistic sourcing and provides a valuable tool for researchers, educators, and consumers seeking to understand the mechanisms of media polarization.

\section*{Limitations}
While this study provides a robust, large-scale analysis of epistemic strategies, its findings should be considered in light of several limitations that also point toward avenues for future research.

First, the scope of our corpus, while large, has boundaries. Our analysis is confined to two major US-based partisan outlets, CNN and Fox News. This binary comparison illuminates the strategies at two poles of the mainstream media spectrum but does not capture the practices of centrist outlets (e.g., Associated Press, Reuters), non-US media, or the rapidly growing ecosystem of hyper-partisan and independent digital media. Furthermore, our study focuses exclusively on digital news articles. The epistemic and sourcing strategies employed in television broadcasts, podcasts, or social media posts from these same outlets may differ due to the distinct affordances and constraints of those mediums.

Second, there are methodological limitations inherent in our analytical approach. Our analysis is conducted at the sentence level, which, while enabling a granular classification of claims, does not capture larger discourse structures. For example, a source may be introduced with full credentials in one sentence but referred to only by pronoun in subsequent sentences; our model would not link these subsequent claims to the original, fully credentialed source. This approach also cannot assess the relative prominence of a sourced claim within an article—a claim in the lead paragraph carries different weight than one in the final paragraph. Additionally, our article-matching strategy, which relies on title similarity, is a powerful heuristic for controlling for topic selection but is not infallible. While the method has extremely high precision (with all manually reviewed matched pairs in the high similarity group in fact reporting on the same event or story) its recall might be lower if articles covering the same event or story happen to have very different titles.

Third, our analysis relies on automated classification. While the underlying model is robust and achieves a relatively high macro-$F_1$ score, it is nevertheless imperfect, which introduces a degree of noise into the aggregate results. 

More broadly, our study demonstrates a strong correlation between media outlets and specific epistemic strategies, but it cannot determine causality or intent. We can only speculate whether these divergent patterns are conscious editorial choices aimed at persuasion or the result of ingrained, differing journalistic cultures. Similarly, while we analyze the textual construction of authority, we do not measure the downstream effects on audience trust, belief, or polarization. Exploring these audience-level effects remains a critical area for future research, which could combine our computational methods with surveys and experimental designs.

\bibliography{custom}

\newpage

\appendix

\begin{table*}[!ht]
\section{Matching Examples}
    \centering
    \footnotesize 
    \begin{tabularx}{\textwidth}{ >{\RaggedRight}X >{\RaggedRight}X c }
        \toprule
        \textbf{Fox News Title} & \textbf{CNN Title} & \textbf{Similarity} \\
        \midrule
        Pope to visit Iraq in March, first foreign trip since coronavirus pandemic began: Vatican & Pope Francis to visit Iraq in March - CNN & 0.76 \\
        \addlinespace
        Midland, Michigan bracing for record flooding as heavy rainfall breaches dams & Michigan governor orders thousands to evacuate after two dams failed, saying one city could be under '9 feet of water' & 0.66 \\
        \addlinespace
        ACLU files complaint over facial recognition arrest | Fox News & ACLU files complaint claiming the first known case of mistaken arrest based on facial recognition identification - CNN & 0.78 \\
        \addlinespace
        Protesters clash with Minneapolis police after George Floyd’s death & Minneapolis: Hundreds protest in streets after George Floyd died following police encounter - CNN & 0.75 \\
        \addlinespace
        DeSantis Education Dept. puts Broward County School Board on notice: Curb mask mandate or lose paychecks & Florida Gov. Ron DeSantis threatens salaries if schools mandate masks & 0.77 \\
        \addlinespace
        Baltimore restaurant apologizes after black woman, son denied service & Baltimore restaurant Ouzo Bay denied service to a Black boy for his clothes. A White boy, dressed similarly, was allowed - CNN & 0.67 \\
        \addlinespace
        Texas abortion law: Biden's DOJ asks judge to intervene & DOJ asks federal judge to halt enforcement of Texas abortion law & 0.75 \\
        \addlinespace
        A\$AP Rocky says Rihanna is 'the one': 'The love of my life' & A\$AP Rocky says he and Rihanna are dating & 0.79 \\
        \addlinespace
        Wealthy Hamptons party hosts use rapid tests for guests to keep up lavish lifestyle: report & In the Hamptons, some hosts are paying for rapid coronavirus tests for guests - CNN & 0.76 \\
        \addlinespace
        Federal judge mulls contempt as Supreme Court hands trump win on deportations | Fox News & Supreme Court backs Trump for now on fired probationary federal employees | CNN Politics & 0.63 \\
        \addlinespace
        Brooks Koepka withdraws from Travelers Championship after caddie tests positive for coronavirus & Travelers Championship: 5 golfers withdraw from PGA Tour event over potential coronavirus exposures - CNN & 0.71 \\
        \bottomrule
    \end{tabularx}
    \caption{Examples of matched article titles from Fox News and CNN in the high-similarity group.}
    \label{tab:matched_titles}
\end{table*}

\clearpage

\begin{table*}[htbp]
\section{Annotated Article Comparison} 
\centering
\footnotesize 

\resizebox{0.85\textwidth}{!}{\begin{minipage}{\textwidth} 
\begin{tabularx}{\textwidth}{ >{\RaggedRight\arraybackslash}X >{\RaggedRight\arraybackslash}X }
\toprule
\textbf{\textit{Fox News} - Second child with measles has died in Texas, officials say} & \textbf{\textit{CNN} - New York Times: Second child dies of measles in Texas} \\
\midrule

\begin{xmlcode}
\xmlopenA{Fact\_Appeal}{Indirect\_Quote}A second child with measles in Texas has died, although the exact cause of death is unknown at this time,\xmlcloseA{Fact\_Appeal}{Indirect\_Quote} according to the \xmlopenB{Appeal\_Source}{Named}{Official}Department of Health and Human Services\xmlcloseB{Appeal\_Source}{Named}{Official}.

\par\addvspace{\baselineskip}
\xmlopenA{Fact\_Appeal}{Indirect\_Quote}HHS Secretary Robert F. Kennedy, Jr. plans to attend the child’s funeral on Sunday,\xmlcloseA{Fact\_Appeal}{Indirect\_Quote} \xmlopenB{Appeal\_Source}{Unnamed}{Official}a spokesperson\xmlcloseB{Appeal\_Source}{Unnamed}{Official} told \xmlopenN{Recipient}NBC News.\xmlcloseN{Recipient}

\par\addvspace{\baselineskip}
\xmlopenN{Fact\_No\_Appeal}The pair of children and an adult in New Mexico who is also believed to have died from measles are the first reported deaths in connection with the disease in the country in a decade.\xmlcloseN{Fact\_No\_Appeal}

\par\addvspace{\baselineskip}
\xmlopenA{Fact\_Appeal}{Indirect\_Quote}MEASLES OUTBREAK CONTINUES: SEE WHICH STATES HAVE REPORTED CASESSince January, 481 cases of measles have been confirmed in Texas alone ,\xmlcloseA{Fact\_Appeal}{Indirect\_Quote} according to the \xmlopenB{Appeal\_Source}{Named}{Official}Texas Department of State Health Services.\xmlcloseB{Appeal\_Source}{Named}{Official}

\par\addvspace{\baselineskip}
\xmlopenA{Fact\_Appeal}{Indirect\_Quote}That total includes six infants and toddlers at a day care center in Lubbock who tested positive within the past two weeks,\xmlcloseA{Fact\_Appeal}{Indirect\_Quote} according to \xmlopenB{Appeal\_Source}{Named}{News\_Report}NBC News.\xmlcloseB{Appeal\_Source}{Named}{News\_Report}

\par\addvspace{\baselineskip}
\xmlopenN{Fact\_No\_Appeal}Two of those children are among 56 people who have been hospitalized with measles in the area since the disease began spreading in January.\xmlcloseN{Fact\_No\_Appeal}

\par\addvspace{\baselineskip}
\xmlopenA{Fact\_Appeal}{Indirect\_Quote}PARENTS OF GIRL WHO DIED AFTER MEASLES INFECTION SAID THEY WOULDN'T GET MMR VACCINE\xmlcloseA{Fact\_Appeal}{Indirect\_Quote}

\par\addvspace{\baselineskip}
\xmlopenA{Fact\_Appeal}{Indirect\_Quote}About one to three out of every 1,000 children infected with measles die from respiratory and neurological complications,\xmlcloseA{Fact\_Appeal}{Indirect\_Quote} according to \xmlopenB{Appeal\_Source}{Named}{Expert\_Document}data from the Centers for Disease Control and Prevention\xmlcloseB{Appeal\_Source}{Named}{Expert\_Document}.

\par\addvspace{\baselineskip}
\xmlopenN{Fact\_No\_Appeal}About one out of every 20 children with measles suffers from pneumonia , which is the most common cause of death from measles in young children.\xmlcloseN{Fact\_No\_Appeal}

\par\addvspace{\baselineskip}
\xmlopenN{Fact\_No\_Appeal}The measles outbreak began in Texas in late January but has since spread to a few other states.\xmlcloseN{Fact\_No\_Appeal}

\par\addvspace{\baselineskip}
\xmlopenA{Fact\_Appeal}{Indirect\_Quote}In the U.S., 628 measles cases have been reported in at least 21 states and Washington, D.C., so far this year,\xmlcloseA{Fact\_Appeal}{Indirect\_Quote} according to \xmlopenB{Appeal\_Source}{Named}{News\_Report}NBC News\xmlcloseB{Appeal\_Source}{Named}{News\_Report}.
\end{xmlcode}
&

\begin{xmlcode}
\xmlopenA{Fact\_Appeal}{Indirect\_Quote}An 8-year-old girl in Texas died Thursday morning of ''measles pulmonary failure,''\xmlcloseA{Fact\_Appeal}{Indirect\_Quote} according to \xmlopenB{Appeal\_Source}{Named}{News\_Report}The New York Times\xmlcloseB{Appeal\_Source}{Named}{News\_Report}, citing \xmlopenB{Appeal\_Source}{Unnamed}{Direct\_Evidence}records it obtained.\xmlcloseB{Appeal\_Source}{Unnamed}{Direct\_Evidence}

\par\addvspace{\baselineskip}
\xmlopenB{Appeal\_Source}{Named}{Official}A Trump administration official\xmlcloseB{Appeal\_Source}{Named}{Official} told \xmlopenN{Recipient}the paper\xmlcloseN{Recipient} \xmlopenA{Fact\_Appeal}{Direct\_Quote}the girl’s cause of death is ''still being looked at.''\xmlcloseA{Fact\_Appeal}{Direct\_Quote}

\par\addvspace{\baselineskip}
\xmlopenN{Fact\_No\_Appeal}This is the second death in the state linked to the ongoing measles outbreak.\xmlcloseN{Fact\_No\_Appeal}

\par\addvspace{\baselineskip}
\xmlopenN{Fact\_No\_Appeal}The first death was in an unvaccinated school-age child in February.\xmlcloseN{Fact\_No\_Appeal}

\par\addvspace{\baselineskip}
\xmlopenN{Fact\_No\_Appeal}A death in New Mexico remains under investigation.\xmlcloseN{Fact\_No\_Appeal}

\par\addvspace{\baselineskip}
\xmlopenA{Fact\_Appeal}{Indirect\_Quote}The outbreak - now spanning Texas, New Mexico, Oklahoma and possibly Kansas - reached at least 569 cases Friday,\xmlcloseA{Fact\_Appeal}{Indirect\_Quote} according to \xmlopenB{Appeal\_Source}{Unnamed}{Expert\_Document}data obtained from state health departments.\xmlcloseB{Appeal\_Source}{Unnamed}{Expert\_Document}

\par\addvspace{\baselineskip}
\xmlopenN{Fact\_No\_Appeal}Texas has reported 481 outbreak-associated cases; New Mexico has 54 cases, and Oklahoma reported 10 cases - eight confirmed and two probable - as of Friday.\xmlcloseN{Fact\_No\_Appeal}

\par\addvspace{\baselineskip}
\xmlopenA{Fact\_Appeal}{Indirect\_Quote}Cases in Kansas, which \xmlopenB{Appeal\_Source}{Unnamed}{Official}the state health department\xmlcloseB{Appeal\_Source}{Unnamed}{Official} said may be linked to the outbreak, reached 24 as of Wednesday.\xmlcloseA{Fact\_Appeal}{Indirect\_Quote}

\par\addvspace{\baselineskip}
\xmlopenB{Appeal\_Source}{Unnamed}{Expert}Experts\xmlcloseB{Appeal\_Source}{Unnamed}{Expert} said \xmlopenA{Fact\_Appeal}{Indirect\_Quote}these numbers are most likely a severe undercount because many cases go unreported.\xmlcloseA{Fact\_Appeal}{Indirect\_Quote}

\par\addvspace{\baselineskip}
\xmlopenN{Fact\_No\_Appeal}Most of the reported cases are in people under 18,\xmlcloseN{Fact\_No\_Appeal} and \xmlopenB{Appeal\_Source}{Unnamed}{Expert}experts\xmlcloseB{Appeal\_Source}{Unnamed}{Expert} worry about \xmlopenA{Fact\_Appeal}{Indirect\_Quote}increasing hospitalizations, especially in younger children who are at higher risk of complications.\xmlcloseA{Fact\_Appeal}{Indirect\_Quote}

\par\addvspace{\baselineskip}
\xmlopenA{Fact\_Appeal}{Direct\_Quote}''The more children who get the disease means that there’s an increased chance that there will be more children getting sicker with complications from measles,”\xmlcloseA{Fact\_Appeal}{Direct\_Quote} said \xmlopenB{Appeal\_Source}{Named}{Expert}Dr. Christina Johns\xmlcloseB{Appeal\_Source}{Named}{Expert}, \xmlopenN{Source\_Attribute}a pediatric emergency physician at PM Pediatrics in Annapolis, Maryland.\xmlcloseN{Source\_Attribute}
\end{xmlcode}
\\
\bottomrule
\end{tabularx}
\end{minipage}} 
\caption{A side-by-side comparison of two news articles covering the same event, fully annotated with the \textsc{FactAppeal} classifier.}
\label{tab:annotated_comparison}
\end{table*}

\clearpage

\begin{figure*}[!h]
\section{Additional Results}

\centering
\def\svgwidth{1.0\textwidth}
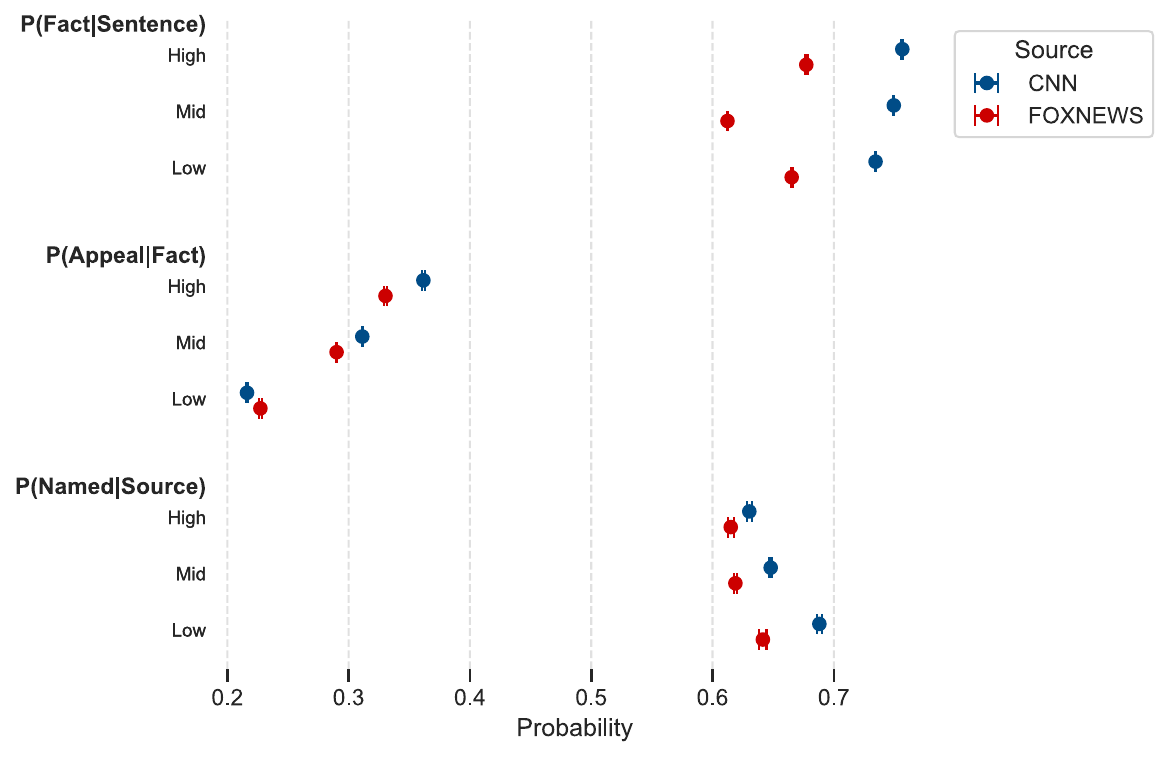
\caption{Key Probabilities by Similarity Group.}
\label{fig:key_probs_group}
\end{figure*}

\begin{figure*}[!h]
\centering
\def\svgwidth{1.0\textwidth}
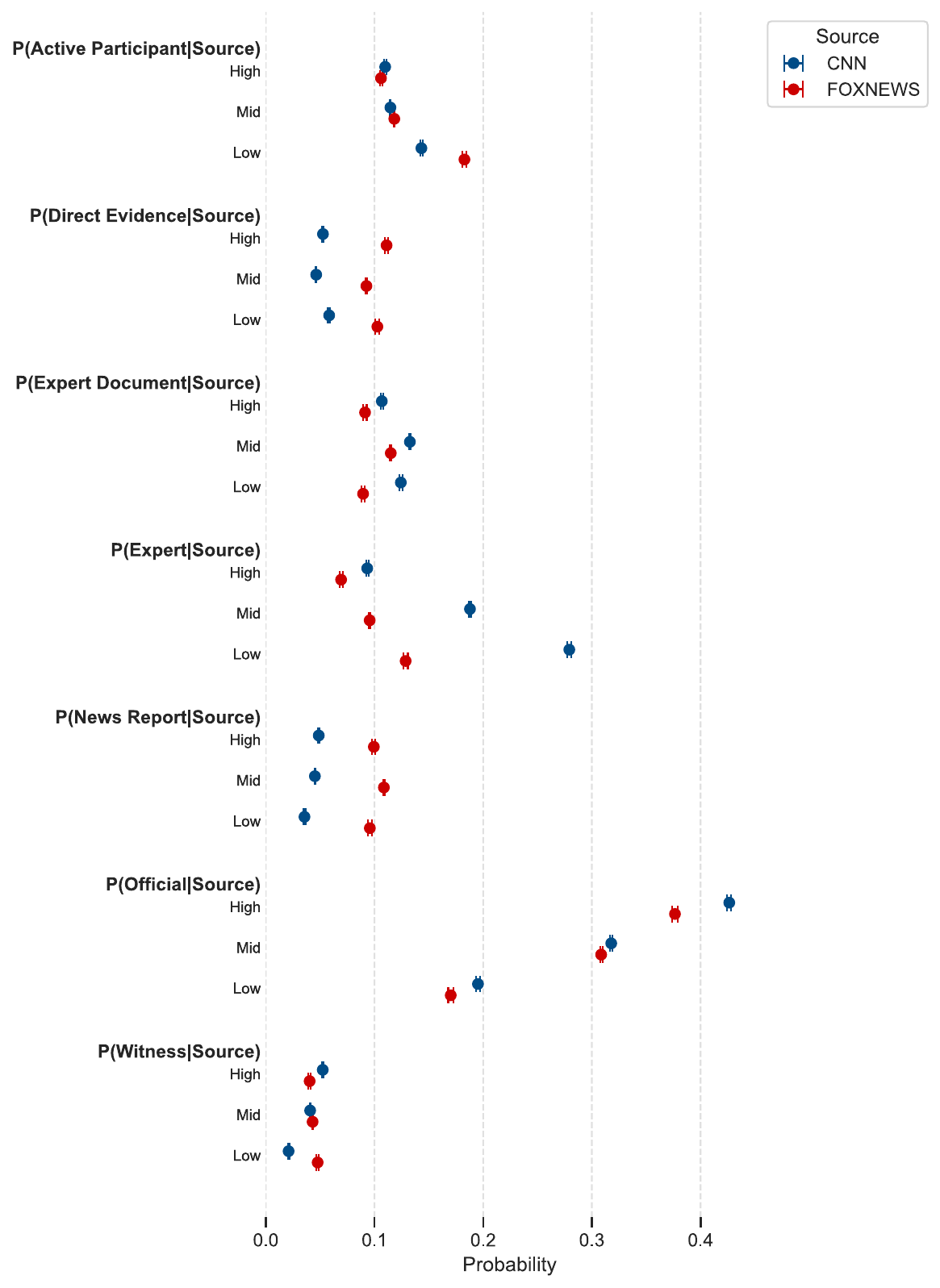
\caption{Source Type Probabilities by Similarity Group.}
\label{fig:source_types_group}
\end{figure*}

\begin{figure*}[!h]
\centering
\def\svgwidth{1.0\textwidth}
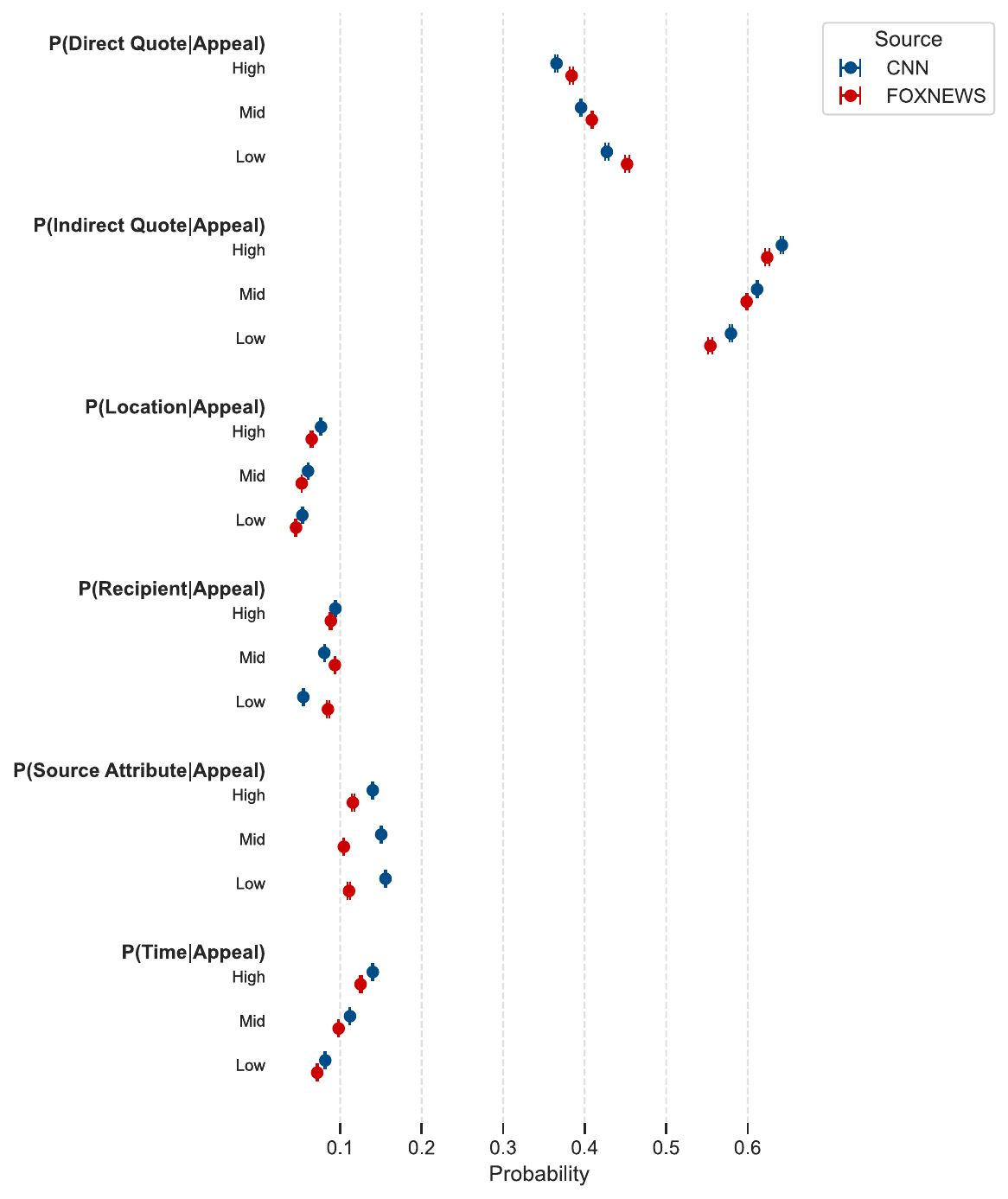
\caption{Appeal Feature Probabilities by Similarity Group.}
\label{fig:appeal_features_group}
\end{figure*}

\clearpage

\begin{table*}[htbp]
\section{Topic Model Details}
\label{sec:appendix}
\centering
\caption{LDA Topic Model Results with Top Terms}
\label{tab:lda_topics}
\resizebox{\textwidth}{!}{%
\begin{tabularx}{1.2\textwidth}{l >{\RaggedRight\arraybackslash}X}
\toprule
\textbf{Topic Name} & \textbf{Top Terms} \\
\midrule
Legal System, Criminal Justice & court case investig attorney justic judg alleg charg law feder file trial legal suprem prosecutor \\
Int'l Competitions, Demographics & women men race french gold pari franc femal black woman medal london italian red itali \\
Technology and Gadgets & amazon use set appl best origin test sale featur devic pro app home price phone \\
US Politics and Elections & democrat vote elect republican senat trump voter parti campaign hous poll polit candid gop presid \\
Crime, Policing, and Protests & polic offic black shoot citi arrest kill crime death gun protest man murder suspect depart \\
\addlinespace
Russia-Ukraine War & russia ukrain russian secur militari offici forc putin war nation intellig ukrainian defens countri unit \\
Energy, Climate, and Vehicles & energi ga oil power provid electr data climat product plant vehicl network materi car market \\
Digital Media, Social Networks & fox apo click post app digit twitter comment continu told email follow media subscrib imag \\
Civics, Government, Social Movements & protest countri nation american right group govern world america church leader polit support mani histori \\
Travel and Int'l News Events & photo caption hide flight airport march passeng taliban plane februari air cruis ship novel crew \\
\addlinespace
Health and Medicine & health medic studi diseas patient research risk test use care hospit doctor infect viru treatment \\
Education and Family & school student children univers parent child educ kid teacher colleg famili princ harri royal girl \\
Science, Environment, and Space & water research world studi island anim human climat earth chang space univers scientist sea use \\
American Football (NFL) & game season team nfl play sport coach yard footbal bowl super week field player quarterback \\
Online Discourse, Misinformation & video trump feedback media sourc watch claim polit whi critic musk social post opinion theori \\
\addlinespace
US Local News and Weather & counti accord california area storm florida park car vehicl west home sever offici near weather \\
Personal Life, Family, Obituaries & famili home york citi help die live life death work son mother father friend lost \\
COVID-19 Pandemic & coronaviru pandem test mask countri case travel viru outbreak health week wear spread reopen citi \\
General Sports and Competitions & game team play sport player win point season second athlet leagu olymp score final open \\
US Presidency and Administration & trump biden presid hous white american administr joe democrat harri donald nation vice campaign obama \\
\addlinespace
Business, Finance, and Economy & compani china busi market million price chines billion rate month trade economi econom accord stock \\
US Political Issues (Immigration, Health) & vaccin border immigr administr migrant health cuomo dose variant offici week number texa illeg million \\
Entertainment, Film, and Music & star film entertain music movi seri award imag actor perform fan celebr love includ share \\
Government and Corporate Policy & work includ govern use feder fund program compani polici servic need provid plan million employe \\
Travel, Dining, Personal Finance & card food travel credit point purchas offer restaur hotel best cash use holiday earn eat \\
\addlinespace
Lifestyle and Design & design work dog use insid build car train look befor place open hand space room \\
Israeli-Palestinian Conflict, Geopolitics & israel attack hama isra iran gaza war palestinian north kill nbsp korea terrorist hostag nuclear \\
\midrule
\multicolumn{2}{l}{\textit{Note: Three junk topics containing technical jargon or conversational filler words were omitted.}} \\
\bottomrule
\end{tabularx}%
}
\end{table*}
\clearpage
\section{Use of Artifacts}
The FactAppeal dataset and model are released under the cc-by-4.0 license. The use of the model in this study is compatible with its intended use.
\end{document}